\documentclass{article}

\usepackage[preprint]{neurips_2025}

\usepackage[utf8]{inputenc} 
\usepackage[T1]{fontenc}    
\usepackage{hyperref}       
\usepackage{url}            
\usepackage{booktabs}       
\usepackage{amsfonts}       
\usepackage{nicefrac}       
\usepackage{microtype}      
\usepackage{xcolor}         
\usepackage{graphicx}
\usepackage{multirow}
\usepackage{enumitem}
\usepackage{pifont}

\def\method{DriveCamSim}

\title{DriveCamSim: Generalizable Camera Simulation via Explicit Camera Modeling for Autonomous Driving}

\author{%
Wenchao Sun\textsuperscript{\textnormal{1,2,3}} \quad
Xuewu Lin\textsuperscript{\textnormal{2}} \quad
Keyu Chen\textsuperscript{\textnormal{1}} \quad
Zixiang Pei\textsuperscript{\textnormal{3}} \AND
Yining Shi\textsuperscript{\textnormal{1}} \quad
Chuang Zhang\textsuperscript{\textnormal{1}} \quad 
Sifa Zheng\textsuperscript{\textnormal{1}}
\vspace{0.2cm}\\
$^1$ Tsinghua University \quad
$^2$ Horizon \quad
$^3$ Horizon Continental Technology
}

\begin{document}
\maketitle
\begin{abstract}
Camera sensor simulation serves as a critical role for autonomous driving (AD), e.g. evaluating vision-based AD algorithms. While existing approaches have leveraged generative models for controllable image/video generation, they remain constrained to generating multi-view video sequences with fixed camera viewpoints and video frequency, significantly limiting their downstream applications. To address this, we present a generalizable camera simulation framework DriveCamSim, whose core innovation lies in the proposed Explicit Camera Modeling (ECM) mechanism. Instead of implicit interaction through vanilla attention, ECM establishes explicit pixel-wise correspondences across multi-view and multi-frame dimensions, decoupling the model from overfitting to the specific camera configurations (intrinsic/extrinsic parameters, number of views) and temporal sampling rates presented in the training data. For controllable generation, we identify the issue of information loss inherent in existing conditional encoding and injection pipelines, proposing an information-preserving control mechanism. This control mechanism not only improves conditional controllability, but also can be extended to be identity-aware to enhance temporal consistency in foreground object rendering. With above designs, our model demonstrates superior performance in both visual quality and controllability, as well as generalization capability across spatial-level (camera parameters variations) and temporal-level (video frame rate variations), enabling flexible user-customizable camera simulation tailored to diverse application scenarios. Code will be avaliable at \url{https://github.com/swc-17/DriveCamSim} for facilitating future research.
\end{abstract}
\section{Introduction}
\label{sec:intro}

\begin{figure}[htbp]
  \centering
  \includegraphics[width=0.9\linewidth]{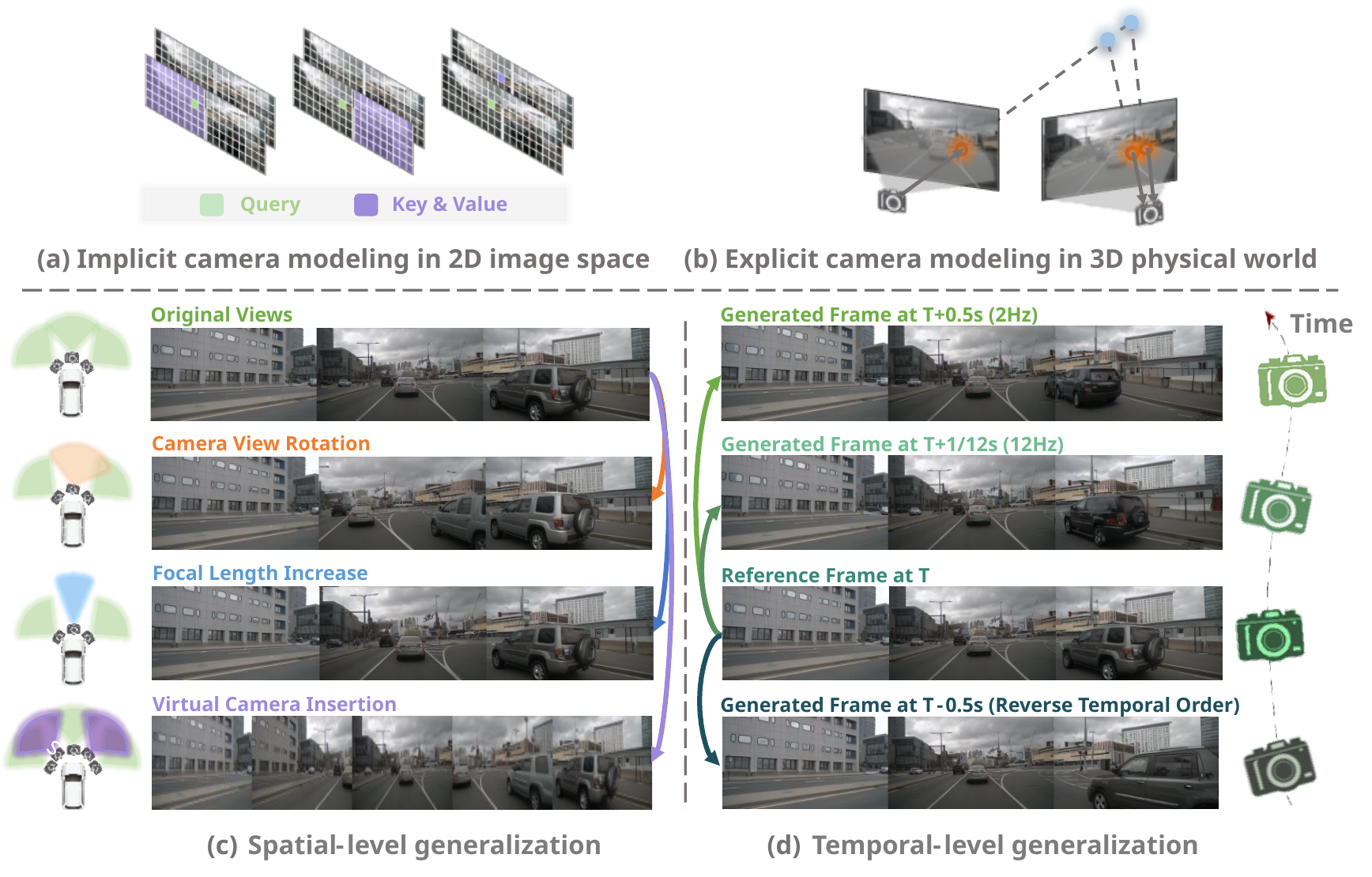}
  \caption{Instead of (a) implicit camera modeling in 2D image space, we propose (b) explicit camera modeling in 3D physical world to unleash the (c) spatial-level and (d) temporal-level generalization capabilities for flexible camera simulation.}
  \label{fig:generalization}
\end{figure}

The field of autonomous driving (AD) has witnessed significant progress in recent years, benefiting from the emergence of large-scale datasets and technological progress. This evolution has propelled the paradigm shift from conventional modular frameworks to integrated end-to-end systems\cite{uniad, vad, sparsedrive, diffusiondrive} and knowledge-enhanced learning methodologies\cite{drivevlm, dilu, senna}. Despite demonstrating impressive performance on standardized benchmarks, critical limitations persist in terms of generalization capability and performance in corner cases. These shortcomings primarily stem from the limited data diversity inherent in existing evaluation frameworks, highlighting the urgent need for more realistic simulation platforms.

To facilitate the development of vision-based AD algorithms, recent studies have employed advanced techniques such as NeRF\cite{nerf}, 3D GS\cite{3dgs}, and diffusion models\cite{ddpm} to synthesize multi-view driving scenes. Among these, diffusion-based approaches have garnered substantial research interest due to their exceptional capability in generating highly realistic and diverse scenarios while offering flexible conditional control.

However, a critical limitation persists in existing approaches: most prior works inherently assume fixed camera parameters and frame rates, which significantly deviates from real-world deployment scenarios. As illustrated in Fig. \ref{fig:generalization}(a), current methods typically employ vanilla attention to model intra-view, cross-view, and cross-frame interactions. This can be seen as an implicit camera modeling in 2D image space that overfits to specific camera parameters and video frequency presented in training dataset, thus exhibit poor generalization capability, severely restricting their practical applications. Although few research\cite{gaia2} have tried to address this issue by augmenting training dataset with different camera rigs and video frequency, the fundamental limitation of generalization beyond the training distribution remains unresolved.

To overcome this limitation, we propose DriveCamSim, a generalizable camera simulation framework with the core lying in Explicit Camera Modeling (ECM) as shown in Fig. \ref{fig:generalization} (b). Leveraging the 3D physical world as a bridge, ECM builds explicit pixel-wise correspondence across multi-view and multi-frame. This approach decouples the model from overfitting to specific camera parameters for multi-view and breaks the chronological order for multi-frame, thus unleashing the generalization capability across spatial-level (varying intrinsic/extrinsic parameters, number of views, Fig. \ref{fig:generalization} (c)) and temporal-level (different video frequency, Fig. \ref{fig:generalization} (d)), even trained on dataset lacking such diversity. Building on ECM’s strengths, we further introduce an overlap-based view matching strategy to dynamically select the most relevant context, and a random frame sampling strategy to mitigate the issue of over-reliance on temporal adjacent frames during generation.

For controllable generation, we identify the issue of information loss inherent in existing conditional encoding and injection pipelines, as shown in Fig. \ref{fig:control_mechanism} (a) and (b), and propose an information-preserving control mechanism to alleviate this issue. Furthermore, our control mechanism can be extended to be identity-aware with foreground appearance features from reference frames, yielding better controllability and foreground temporal consistency.

To summary, our contributions as summarized as follows:
\begin{itemize}[leftmargin=*]
\item  We propose DriveCamSim, a novel generalizable camera simulation framework with the core idea of Explicit Camera Modeling, along with an overlap-based view matching and a random frame sampling strategy. These designs not only enhance visual quality, but also unleash the generalization capability across spatial-level and temporal-level, supporting flexible camera simulation for downstream application.
\item  We diagnose and address critical information loss in existing conditional pipelines, proposing an information-preserving control mechanism for better controllability, which can be extended to be identity-aware to enhance foreground temporal consistency.
\item  Through extensive experiments, we demonstrate state-of-the-art performance in visual quality, controllability and generalization capability, with ablation studies validating the efficacy of our key designs.
\end{itemize}
\section{Related Works}
\label{sec:related_works}

\subsection{Cross-View Interaction for Multi-View Image Generation}
Effective cross-view interaction is crucial for maintaining spatial consistency in overlapping regions between adjacent camera views. Existing approaches predominantly employ multi-head attention for cross-view modeling, where image patches from one view serve as queries while patches from neighboring views provide keys and values\cite{bevcontrol, panacea}. Recent advancements include MagicDrive\cite{magicdrive}, which incorporates camera parameters as scene-level conditioning, and DriveDreamer-2\cite{drivedreamer2}, which reformulates cross-view interaction as intra-view processing by concatenating multi-view images along the width dimension.
However, these methods inherently assume fixed camera configurations during training, leading to model specialization on specific viewpoint geometries. This fundamental limitation results in constrained generation capability that cannot extrapolate beyond the trained camera parameter distribution, significantly restricting practical deployment scenarios. In contrast, our framework overcomes this limitation by enabling generalization across diverse camera configurations during inference, thereby supporting flexible camera simulation for real-world applications.

\subsection{Cross-Frame Interaction for Multi-View Video Generation}
Maintaining temporal consistency in video generation requires effective cross-frame interaction. While most existing methods employ multi-head attention to model temporal relationships, they rely on spatially aligned patches in 2D image space, which often fail to maintain alignment in the 3D physical world —particularly in high-speed scenarios. This implicit modeling make the model overfit to the specific video frequency in training dataset, limiting their applicability in real-world settings. For instance, DreamForge\cite{dreamforge} generates 7-frame clips at 12Hz but only utilizes the last frame as input for 2Hz driving agent\cite{drivearena}, resulting in inefficient computation. Furthermore, while high-frequency training data can be downsampled to produce low-frequency outputs, the reverse is not feasible. In contrast, our approach is able to generalizing across varying frame rates , enabling high-frequency generation from low-frequency training data, and even support generation in reverse temporal order.

\subsection{Control Mechanism for 3D Condition}
\begin{figure}[htbp]
  \centering
  \includegraphics[width=0.9\linewidth]{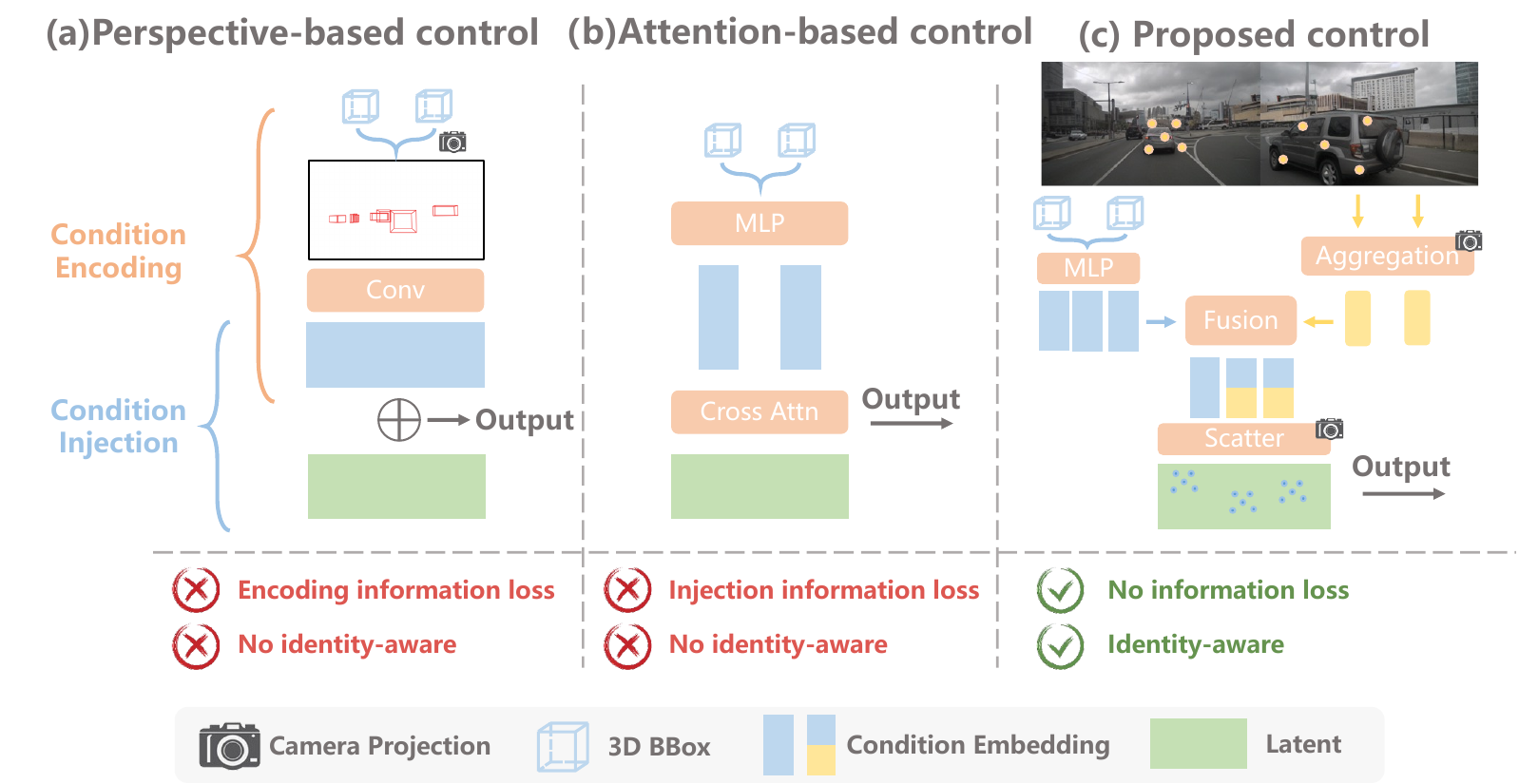}
  \caption{Our control mechanism preserves information in encoding and injection stage, and support identity feature encoding.}
  \label{fig:control_mechanism}
\end{figure}

The control mechanism operates through two sequential stages: (1) the condition encoding stage transforms low-dimensional control signals into high-dimensional condition embeddings, and (2) the condition injection stage incorporates these embeddings into the image latent space. Current approaches can be categorized into two predominant paradigms:
\textbf{Perspective-based Control}\cite{drivedreamer, panacea}: As shown in Fig. \ref{fig:control_mechanism} (a), this method projects 3D bounding boxes and road layouts onto 2D perspective views during encoding, followed by direct addition to image latents. However, the 3D-to-2D projection inherently suffers from depth information loss. For instance, a large vehicle at a far distance and a small vehicle at close range may produce similarly sized 2D bounding boxes, introducing ambiguity for model learning.
\textbf{Attention-based Control}\cite{magicdrive}: As shown in Fig. \ref{fig:control_mechanism} (b), this approach encodes bounding boxes as instance-level embeddings and integrates them via cross-attention mechanisms. While effective in some scenarios, this paradigm learns implicit view transformations that tend to overfit to specific camera parameters, consequently losing critical relative pose information between objects and the camera.
In contrast, our proposed control mechanism systematically preserves spatial and geometric information throughout both encoding and injection stages.
\begin{figure}[htbp]
  \centering
  \includegraphics[width=\linewidth]{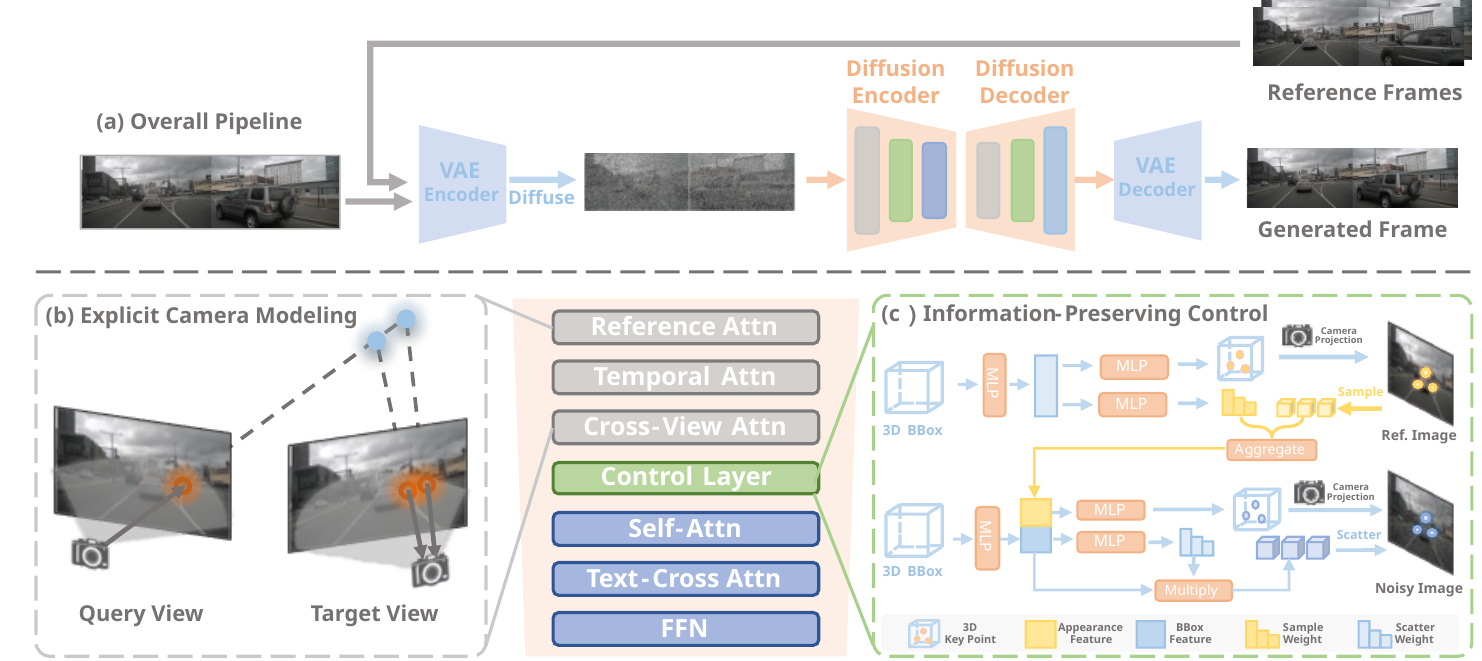}
  \caption{Overall framework of DriveCamSim. The (a) proposed method is built upon a pretrained latent diffusion model\cite{ldm}, with several (b) attention layers and (c) control layers inserted.}
  \label{fig:overall_framework}
\end{figure}

\section{Methods}
\label{sec:method}

\subsection{Problem Formulation}
This work addresses the problem of controllable camera simulation for autonomous driving. Following Bench2Drive-R\cite{bench2drive-r}, at a given time $t$, below information is provided as the input of our generative model:
\begin{enumerate}[leftmargin=*]
\item \textbf{3D bounding boxes}: $\mathbf{B}_t = \{(b_i, c_i)\}_{i=1}^{N_b}$, where $b_i = {(x_i, y_i, z_i, l_i, w_i, h_i, yaw_i)} $ is the bounding boxes for foreground objects (vehicles, pedestrians, bicycles, etc.) within a specific range; $c_i\in \mathcal{C}_{box}$ is the semantic label. 
\item \textbf{Vectorized map elements}: $\mathbf{M}_t = \{(v_i, c_i)\}_{i=1}^{N_m}$, where $v_i = {(x_j, y_j)}_{j=1}^{N_v}$ represents vertices for polygon map elements (cross-walk regions, etc.) and interior points for linestring map elements (road boundaries, lane dividers, etc.); $c_i \in \mathcal{C}_{map}$ represents the map class. 
\item \textbf{Ego pose}: $\mathbf{E}_t \in \mathbb{R}^{4 \times 4}$ is ego pose matrix including ego-to-global translation and rotation.
\item \textbf{Camera parameters}: $\mathbf{K} = \{\mathbf{K}_i\in \mathbb{R}^{4\times 4}\}_{i=1}^{N_{cam}}$, where $\mathbf{K}_i$ is the camera transformation matrix composed of intrinsic and extrinsic matrices that transforms points from ego coordinate system to image coordinate system.
\item \textbf{Reference information}:  
$\mathbf{H}_r = \{(\mathbf{I}_r, \mathbf{K}_r, \mathbf{E}_r, \mathbf{B}_r)\}_{r=1}^{N_r}$, where $N_r$ is the number of reference frames. The reference information includes original recorded images, camera parameters, corresponding pose and boxes, which are used to retrieve box appearance feature.
\item \textbf{Historical information}:  
$\mathbf{H}_h = \{(\mathbf{I}_h, \mathbf{K}_h, \mathbf{E}_h, \mathbf{B}_h)\}_{h=1}^{N_h}$, where $N_h$ is the number of historical frames. $\mathbf{H}_h$ is similar to $\mathbf{H}_r$, except that the reference images ${I}_r$ are logged real images, while historical images ${I}_h$ are previous generated images. 
\end{enumerate}
With these information, out model generate multi-view images at time $t$: $I_t = \mathcal{G}(\mathbf{B}_t, \mathbf{M}_t, \mathbf{E}_t, \mathbf{K}, \mathbf{H}_r, \mathbf{H}_h)$, which will be used as historical information for auto-regressive generation. We adopt such an online generation scheme rather than offline long video generation to enable reactive simulation for downstream AD algorithms.

\subsection{Overall Framework}
The overall framework of DriveCamSim is shown in Fig. \ref{fig:overall_framework}. Our model builds upon a pretrained latent diffusion model\cite{ldm}, with several attention layers and control layers inserted within attention blocks. 

\begin{figure}[htbp]
  \centering
  \includegraphics[width=0.7\linewidth]{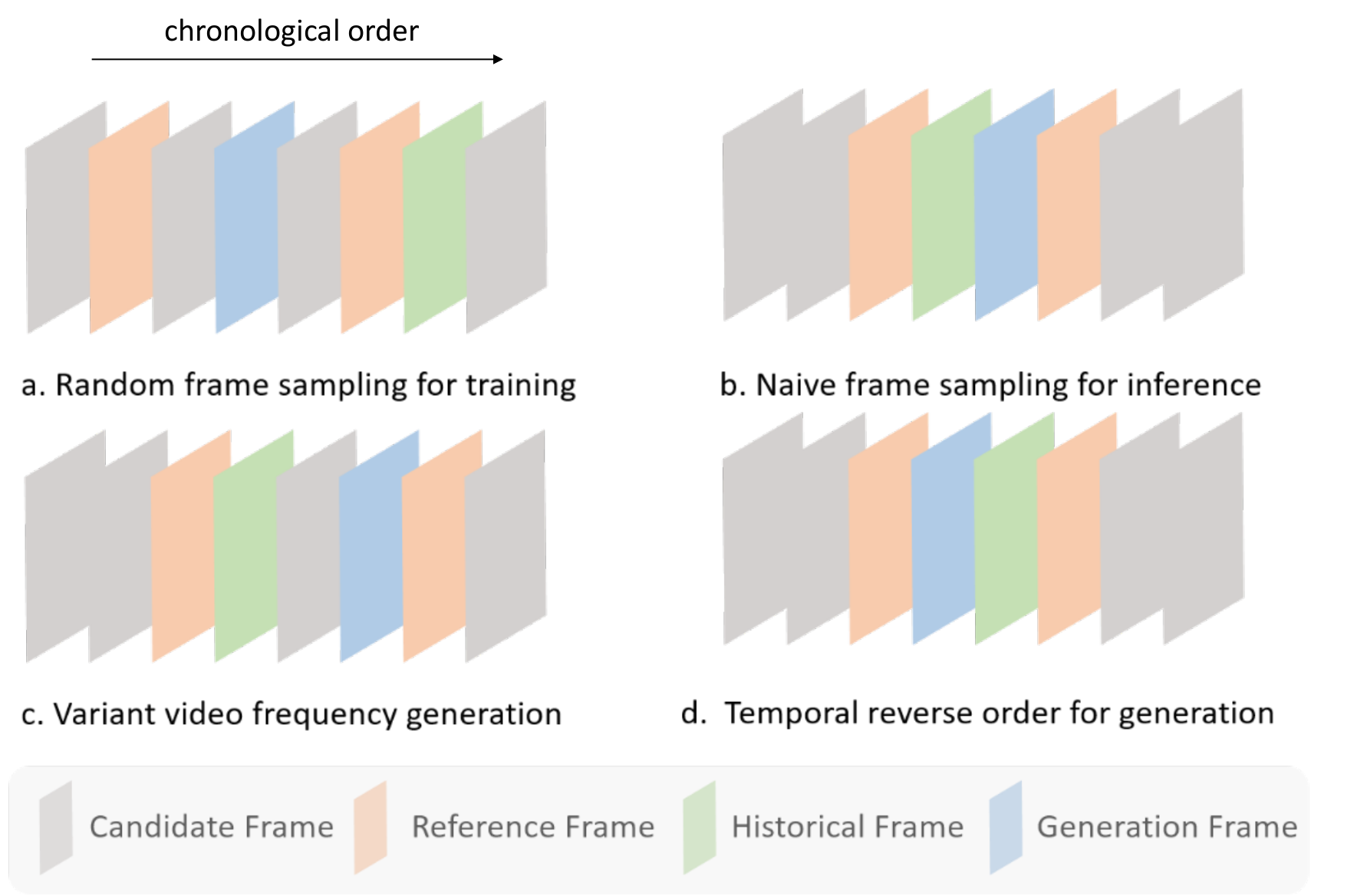}
  \caption{Frame sampling strategy for training and inference.}
  \label{fig:frame_sampling}
\end{figure}

\subsection{Explicit Camera Modeling}
The motivation for Explicit Camera Modeling is to build correspondence between pixels across multi-view and multi-frame, enabling interaction in 3D physical world rather than 2D image space. For simplicity, we take a query view \(V_{query}\) and a target view \(V_{key}\) to illustrate ECM, but can easily be extended to multi views. Query view is selected from current frame, while target view can be selected from current frame (for cross-view attention), reference frame (for reference attention) or historical frame (for temporal attention).

\textbf{Building Pixel Correspondence.}
For each pixel $p_q = (u_q, v_q)$ in \(V_{query}\), we first project it to 3D space. However, regressing to a precise depth value is difficult, especially for noisy latents. So we set several depth anchors $d = \{d_i\}_{i=1}^D$ and back project $p_q$ to 3D points $\{P_{qi}\}_{i=1}^D$, where $P_{qi} = d_i \cdot K^{-1} \cdot p_q$. $\{P_{qi}\}$ are further projected to \(V_{key}\) to get $\{p_{ki}\}_{i=1}^D$, where $p_{ki} = K_k \cdot E_k^{-1} \cdot E_t \cdot P_{qi}$, $K_k$ and $E_k$ are camera projection matrix and global pose of target view. By doing so, we build correspondence between query view pixel $p_q$ and target view pixels $\{p_{ki}\}$.

\textbf{Feature Aggregation.}
After building pixel correspondence, we aggregate features at $\{p_{ki}\}$ to refine query feature at $p_q$. For each $p_q$, we have \(d \) target pixels, considering not all target pixels are equally important, we predict a depth distribution to model the attention weights between $p_q$ and $\{p_{ki}\}$ with $W_{qk} = \mathrm{Softmax}(\mathrm{MLP}(f_q)) \in \mathbb{R}^{d} $, where $f_q = x_q(u, v)$ is the query pixel feature and $x_q$ is the feature of query view. We also note that 3D points $\{P_{qi}\}$ may project outside of \(V_{key}\), so we filter out these outlier points by setting corresponding weights to zero. Then we conduct image interaction by updating query feature with \(f_q=f_q+\sum_{i=1}^D(W_{qki} \times f_{ki})\), where $f_{ki}$ is target pixel feature at $p_{ki}$.

\textbf{Overlap-based Target View Matching.}
Now for query view \(V_{query} = V_{n, t}\) where $n \in \{1, ..., N_{cam}\}$ is index of view, we extend the target view number to more than one. One problem raises that: how to choose the target view? One naive strategy is to choose $\{V_{n-1,t}, V_{n+1,t} \}$ for cross-view attention, $\{V_{n,r}\}$ for reference attention, and $\{V_{n,h}\}$ for temporal attention. However, in scenarios like turning at intersection, $\{V_{n,r}\}$ and $\{V_{n,h}\}$ might have a small overlap with $V_{n, t}$, resulting in invalid computation. To address this, we propose an overlap-based target view matching strategy to dynamically search best target views. We notice that the ineffective computation comes from much zero weights for outlier points, so we use the percentage of $\{p_{ki}\}$ that hit on target view to represent the degree of overlap, and select views that have maximum overlap with query view as target views. This strategy benefit the feature interaction by providing most relevant context from target views.  

\textbf{Frame Sampling.}
Another problem arise that how to sample reference frame and historical frame. One naive method is to sample frames following chronological order. However, we found these adjacent frames share similar context, resulting in over-reliance on adjacent frames when generating current frame. As shown in Fig. \ref{fig:frame_sampling} (a), built upon explicit camera modeling, our model breaks chronological order in multi-frame video, enabling a random sampling strategy at training to force the model learn the geometric transformation from historical and reference frames to generation frame, rather than simply copy the pattern. This training strategy also unleash flexible inference schemes in Fig. \ref{fig:frame_sampling} (b-d), e.g. generation with variant video frequency or temporal reverse order. 

\subsection{Conditional Mechanism}

\textbf{Text Condition.} Following common practice, our model uses text description for scene-level control. We build a simple prompt template as "A driving scene image. \{Weather\}. \{Daytime\}." The prompt is embedded with CLIP text encoder and injected into image latent through text cross attention.
 
\textbf{3D Bonding Boxes Encoding.}
To prevent information loss in 3D-to-2D projection, we directly encode boxes and class label into an instance-level embedding:
 \[ E_{box_i} = \mathrm{MLP}(b_i) + \mathrm{CLIP}(c_i) \]

\textbf{Extension to be Identity-Aware.}
Previous methods only encode geometric information from $b_i$ and semantic information from $c_i$, lacking identity information and let the model learn to match foreground objects from different frames. This may be confusing in some cases, e.g. crowded scenes. To be completely controllable, we additionally encode appearance feature from historical and reference frames. Following sparse-centric perception model\cite{sparse4d}, we similarly encode the box $b_h$ with same identity at historical frame (or reference frame ), then use the box embedding $E_{box_h}$ to generate several keypoints $\{P_j\}_{j=1}^{N_j}$ around the box and corresponding attention weights $\{w_j\}_{j=1}^{N_j}$, then project $\{P_j\}$ to historical frame with $p_j=K_j \cdot P_j$ to sample the feature $\{f_{p_j}\}$, and aggregate appearance feature as $A_{box_h}=\sum_{j=1}^{N_j}w_j \cdot f_{p_j}$. The box embedding is then updated as: 
 \[ E_{box_i} = \mathrm{MLP(b_i)} + \mathrm{CLIP}(c_i) + \sum_{h=1}^{N_h}A_{box_h} + \sum_{r=1}^{N_r}A_{box_r} \]
 
\textbf{Scatter-based Condition Injection.}
To be compatible to our ECM and generalize across different camera parameters, we need to project the condition embedding onto image latent using camera parameters. However, the instance-level embedding is not suitable to directly add to image latents. To address this, we propose a scatter-based condition injection method, which can be regarded as an inverse operation of aggregation. Specifically, we use the condition embedding $E_{box_i}$ to predict several keypoints $\{P_m\}_{m=1}^{N_m}$  around the 3D box $b_i$, and corresponding weights $\{w_m\}_{m=1}^{N_m}$ for each point, the keypoints are projected to image with $p_m=(u_m, v_m)=K_t \cdot P_m$ to find the location on image latent, then the condition embedding is scaled by weights and scatter back to image latent $x$ with $x(u_m, v_m) = x(u_m, v_m) + w_m \cdot E_{box_i} $. In practice, $(u_m, v_m)$ are not integers, so we use bilinear scatter similar to bilinear sampling. 

\subsubsection{Vectorized Map Elements.}
Vectorized map elements are encoded similarly to boxes, and our scatter-based method also applies to map condition injection. 
 \[ E_{vec_i} = \mathrm{MLP}(m_i) + \mathrm{CLIP}(c_i) \]

\begin{table*}[tb!]
\centering
\caption{Comparisons of realism and controllability on nuScenes validation set. * means using real images as reference.}
\resizebox{0.75\linewidth}{!}{
\begin{tabular}{l|c|cccc|cc}
\hline \toprule
\multirow{2}{*}{Method}   & \multirow{2}{*}{FID}  & \multicolumn{4}{c|}{ BEVFusion~\cite{bevfusion} (Camera Branch)}  & \multicolumn{2}{c}{StreamPETR~\cite{streampetr}}  \\ 
&  & NDS$\uparrow$ & mAP$\uparrow$ & mAOE$\downarrow$ & mIoU $\uparrow$ &NDS$\uparrow$ & mAP$\uparrow$ \\
\midrule
Oracle &- &  41.20 & 35.53 & 0.56 & 57.09 &57.10 &48.20 \\
\midrule
BEVControl~\cite{bevcontrol}& 24.85 & - & - & - & -	 & - & -	\\
MagicDrive~\cite{magicdrive} & 16.20 & 23.35 	&12.54 	& 0.77	&28.94 & 35.51 & 21.41\\
Panacea~\cite{panacea} & 16.69 & -&  -& -& -& 32.10 & - \\
Panacea+~\cite{panacea+} & 15.50 &  -& -& -& -& 34.60 & - \\
\method & \textbf{14.07}&  \textbf{23.87} & \textbf{12.75}  & \textbf{0.64} & \textbf{34.84} & \textbf{39.49} & \textbf{22.41} \\
\midrule
Bench2Drive-R$^*$\cite{bench2drive-r} & 10.95 & 25.75 & 13.53  & 0.73 & 42.75 & 40.23 & 24.04
\\ 
\method$^*$ & \textbf{7.86}& \textbf{26.55} & \textbf{14.47}  & \textbf{0.67} & \textbf{43.36} & \textbf{44.16} & \textbf{28.16} \\
\bottomrule \hline
\end{tabular}
}
\label{main_result}
\end{table*}
\begin{table*}[tb!]
\centering
\caption{Performance of UniAD\cite{uniad}'s different tasks on nuScenes validation set. * means using real images as reference.} 
\resizebox{\textwidth}{!}{
\begin{tabular}{l|cc|cccc|cc|c}
\hline \toprule
\multirow{2}{*}{Method}   & \multicolumn{2}{c|}{Detection}  & \multicolumn{4}{c|}{BEV Segmentation}  & \multicolumn{2}{c|}{Planning} & Occupancy  \\ 
& NDS$\uparrow$ & mAP$\uparrow$ & Lanes$\uparrow$ &Drivable$\uparrow$ & Divider$\uparrow$ & Crossing$\uparrow$ & avg.L2(m)$\downarrow$& avg.Col.(\%)$\downarrow$ & mIoU$\uparrow$ \\
\midrule
Oracle & 49.85 & 37.98 & 31.31  & 69.14  & 25.93  & 14.36 &  1.05 &  0.29  & 63.7 \\ 
\midrule
MagicDrive~\cite{magicdrive} &	29.35 & 14.09&	23.73&	55.28&	18.83&	6.57&1.18 & \textbf{0.33}	& 54.6 \\  
\method & \textbf{31.55} & \textbf{14.70}&	\textbf{25.86}&	\textbf{56.44}&	\textbf{20.66}&	\textbf{8.50}&	\textbf{1.16}&	0.40&	\textbf{55.7} \\ 
\midrule
Bench2Drive-R$^*$\cite{bench2drive-r}&	33.04 & 15.16&	25.50 &	56.53 &	\textbf{21.27}&	 8.67&	\textbf{1.15}&	\textbf{0.31}&	55.5 \\ 
\method$^*$ & \textbf{34.88} & \textbf{16.90}&	\textbf{26.31}&	\textbf{58.58}&	21.25 &	\textbf{9.16}&	\textbf{1.15}&	0.40 &	\textbf{57.0} \\ 
\bottomrule \hline
\end{tabular}
}
\label{tab:uniad_result}
\end{table*}

\section{Experiments}
\label{sec:exp}

\subsection{Experimental Setups}

\textbf{Dataset and Baselines.} We employ nuScenes dataset\cite{nuscenes}, which have 700 street-view scenss for training and 150 for validation with 2Hz annotation. Our baseline models include image generation methods (BEVControl,\cite{bevcontrol},  MagicDrive\cite{magicdrive}), video generation methods (Panacea\cite{panacea}, Panecea+\cite{panacea+}) and simulation-oriented method with real images as reference (Bench2Drive-R\cite{bench2drive-r}). 

\textbf{Evaluation Metrics.} We evaluate the generation realism with Frechet Inception Distance (FID). For controllability, we use BEVFusion\cite{bevfusion} to evaluate foreground object detection and background map segmentation, StreamPETR\cite{streampetr} to evaluate temporal consistency of generated image sequences, and UniAD\cite{uniad} for end-to-end planning.

\textbf{Model Setup.} We utilizes pretrained weights from Stable Diffusion v1.5\cite{ldm}, as we do not have trainable copy from ControlNet\cite{controlnet}, we train all parameters of UNet\cite{unet}. The generation resolution is 224$\times$400, and images are sampled using UniPC\cite{unipc} scheduler for 20 steps with CFG at 2.0.
Through we made a distinction between reference frames and historical frames, our explicit camera modeling can handle them in a unified format, enabling flexible inference mode. We set total frames up to 3 ($N_r+N_f =3$), and use 3 historical frames as input by default. For comparison with Bench2Drive-R\cite{bench2drive-r}, we use 1 historical frame and 2 reference frames within recordings with closest distance.

\subsection{Main Results}
\textbf{Generation Realism and Controllability.}
As show in Tab. \ref{main_result}, our method outperforms baselines in generation realism with a lower FID score, and achieves better controllability on both foreground and background generation. 

\textbf{Temporal Consistency.}
As show in Tab. \ref{main_result}, perception results evaluted by StreamPETR\cite{streampetr} are notably better than the baseline methods, whether with or without reference images. This demonstrates the temporal consistency of our auto-regressive generated image sequences.

\textbf{Generation for End-to-End Planning.}
As show in Tab. \ref{tab:uniad_result}, our method outperforms baselines on nearly all metrics, indicating the potential of our method for driving agent simulation.

\subsection{Ablation Study}
For ablation studies, we use a SOTA end-to-end method SparseDrive\cite{sparsedrive} to evaluate on various tasks. There is an important reason to choose SparseDrive, that the model is trained with random augmentation in image space and 3D space, thus can generalize to perturbations in camera parameters to some extent. Experiments for camera parameter generalization are provided in Appendix due to space limit.

\textbf{Ablation for Camera Modeling.}
As demonstrated in Table. \ref{tab:ablation_ecm}, replacing our explicit camera modeling with implicit camera modeling leads to consistent performance degradation across all evaluation metrics, especially for temporal and reference attention, indicating the importance of aligning in 3D physical world rather than 2D image space.

\textbf{Ablation for Overlap-based View Matching and Random Frame Sampling Strategy.}
As illustrated in Table \ref{tab:ablation_strategy}, the ablation study reveals key observations as follows. When overlap-based view matching (OVM) is utilized during training but disabled at inference (ID-2), a marginal performance degradation occurs in all perception tasks. And complete removal of OVM during both training and inference (ID-3) leads to a more pronounced performance drop, underscoring its importance. The exclusion of random frame sampling during training (ID-4) further adversely affects task performance, suggesting its importance for model learning.

\textbf{Ablation for Control Mechanism.}
As show in Tab. \ref{tab:ablation_control}, compared to ID-2, ID-1 introduces appearance feature and brings improvement on tracking metric, indicating better foreground temporal consistency. ID-3 indicates that it's necessary to preserve 3D information in condition encoding, and ID-4 shows attention-based control suffers from slow convergence for losing view transformation information between boxes and cameras.

\subsection{Qualitative Results}
\begin{figure}[htbp]
  \centering
  \includegraphics[width=\linewidth]{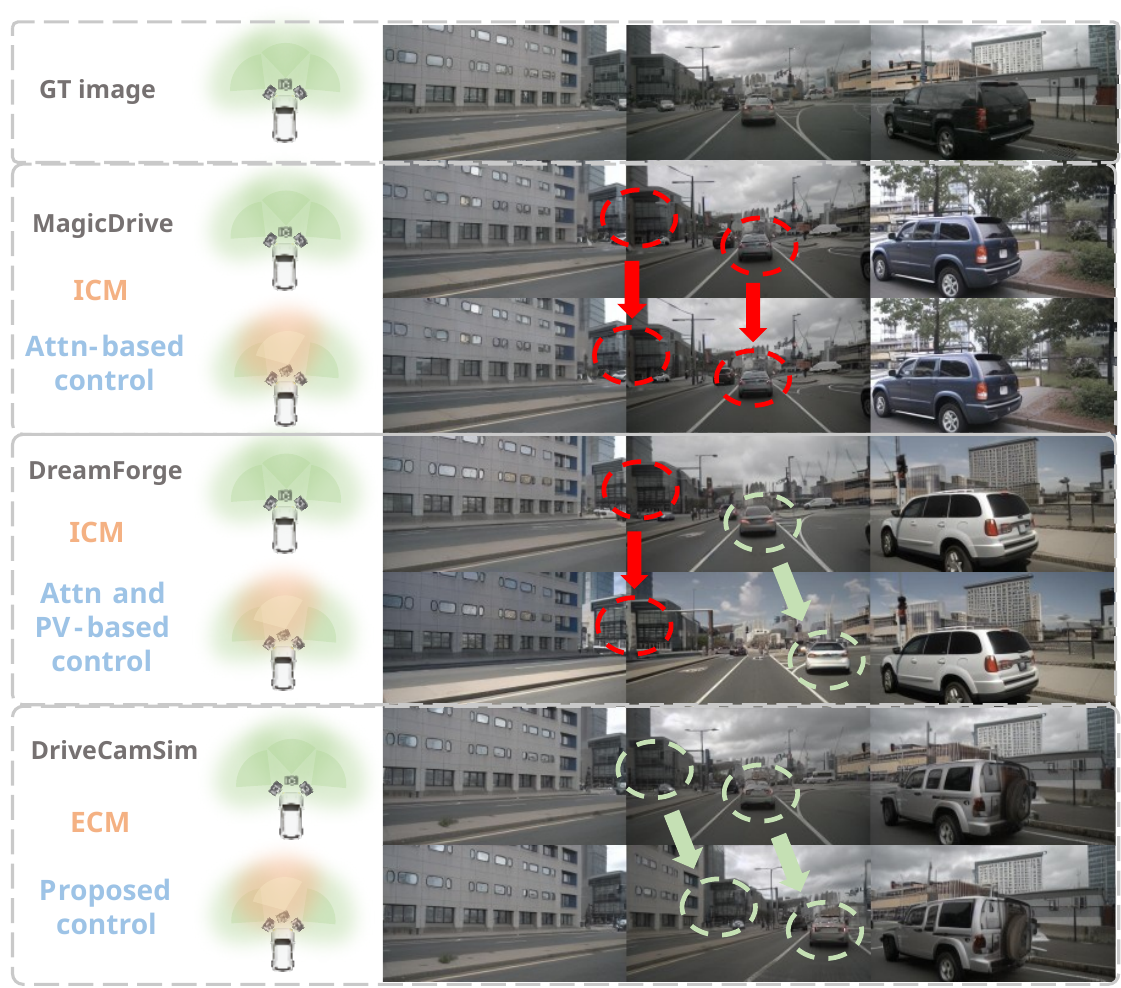}
  \caption{Qualitative results for spatial-level generalization. Rotate front camera 20° to the left, DriveCamSim succeed to generate images with correct foreground and background, while MagicDrive and DreamForge fails.}
  \label{fig:spatial_generalizes}
\end{figure}

We compare our method with MagicDrive\cite{magicdrive} and DreamForge\cite{dreamforge} for spatial-level generalization capability in Fig. \ref{fig:spatial_generalizes}. Taking the example of rotating the front camera 20° to the left, we can find that with implicit camera modeling and attention-based control, MagicDrive generates nearly same images before and after rotation. DreamForge, enhanced with perspective-based control, maintains foreground controllability after rotation, but fails to generate correct background. Our method, with explicit camera modeling and information-preserving control, correctly handles both foreground and background. More visualizetion results are provided in Appendix.

\begin{table}[tb!]
\centering
\caption{Ablation for explicit and implicit camera modeling. ECM-S, ECM-T and ECM-R represent explicit camera modeling for cross-view, cross-frame and reference attention. The implicit camera modeling follows Panacea\cite{panacea}.}
\label{tab:ablation_ecm}
\scriptsize
{    
\begin{tabular}{l|cc|cc|c|c|cc}
\toprule
\multirow{2}{*}{ID} &
\multirow{2}{*}{ECM-S} &
\multirow{2}{*}{ECM-T \& ECM-R} & 
\multicolumn{2}{c|}{3DOD} & 
\multicolumn{1}{c|}{Tracking} & 
\multicolumn{1}{c|}{Online Mapping} &
\multicolumn{2}{c}{Planning} \\
&&& NDS$\uparrow$ & mAP$\uparrow$ & AMOTA$\uparrow$ & mAP$\uparrow$ & L2$\downarrow$ & Col(\%) $\downarrow$ \\
\midrule
1 & \checkmark  & \checkmark & \textbf{36.44} & \textbf{19.57} & \textbf{9.05} & \textbf{22.84} & \textbf{0.69} & \textbf{0.19}  \\
2 &  & \checkmark &  33.83 & 16.01 & 6.76 & 20.26 & 0.79 &  0.28 \\ 
3 & \checkmark &  & 30.67 & 14.84 & 6.08 & 12.95 & 0.83 & 0.36  \\
\bottomrule
\end{tabular}
}
\end{table} 
\begin{table}[tb!]
\centering
\caption{Ablation for overlap-based view matching (OVM) and random frame sampling strategy.}
\label{tab:ablation_strategy}
\scriptsize
{ 
\resizebox{0.9\linewidth}{!}{
\begin{tabular}{l|cc|cc|c|c|cc}
\toprule
\multirow{2}{*}{ID} &
\multirow{2}{*}{OVM at training/infernce} &
\multirow{2}{*}{Random Frame Sampling} & 
\multicolumn{2}{c|}{3DOD} & 
\multicolumn{1}{c|}{Tracking} & 
\multicolumn{1}{c|}{Online Mapping} &
\multicolumn{2}{c}{Planning} \\
&&& NDS$\uparrow$ & mAP$\uparrow$ & AMOTA$\uparrow$ & mAP$\uparrow$ & L2$\downarrow$ & Col(\%)$\downarrow$ \\
\midrule
1 & \checkmark / \checkmark & \checkmark & \textbf{36.44} & \textbf{19.57} & \textbf{9.05} & \textbf{22.84} & \textbf{0.69} & 0.19  \\
2 & \checkmark / \ding{55} & \checkmark & 36.25 & 19.40 & 8.91 & 22.29 & \textbf{0.69} & \textbf{0.18} \\
3 & \ding{55} / \ding{55}  & \checkmark & 33.58 & 16.11 & 6.66 & 18.74 & 0.79 & 0.29  \\
4 & \checkmark / \checkmark &  & 32.90 & 15.24 & 7.00 & 16.95 & 0.72 & 0.28  \\
\bottomrule
\end{tabular}
}
}
\end{table} 
\begin{table}[tb!]
\centering
\caption{Ablation for control mechanism and identity feature.}
\label{tab:ablation_control}
\scriptsize
{    
\begin{tabular}{l|lc|cc|c|c|cc}
\toprule
\multirow{2}{*}{ID} &
\multirow{2}{*}{Control Mechansim} &
\multirow{2}{*}{Identity Aware} &
\multicolumn{2}{c|}{3DOD} & 
\multicolumn{1}{c|}{Tracking} & 
\multicolumn{1}{c|}{Online Mapping} &
\multicolumn{2}{c}{Planning} \\
&&& NDS$\uparrow$ & mAP$\uparrow$ & AMOTA$\uparrow$ & mAP$\uparrow$ & L2$\downarrow$ & Col(\%)$\downarrow$ \\
\midrule
1 & Our Control & \checkmark & \textbf{36.44} & \textbf{19.57} & \textbf{9.05} & \textbf{22.84} & \textbf{0.69} & \textbf{0.19}  \\
2 & Our Control &  & 34.96 & 17.16 & 7.22 & 20.24 & 0.78 & 0.32  \\
3 & Perspective-based control &  & 31.15 & 14.26 & 5.52 & 21.36 & 0.73 & 0.20  \\
4 & Attention-based control & & 26.23 & 10.01 & 4.92 & 15.65 & 0.81 & 0.32  \\
\bottomrule
\end{tabular}
}
\end{table} 
\newpage
\section{Conclusion and Future Work} \label{conclusion}
\label{sec:conclusion}

\paragraph{Conclusion.} In this work, we explore the explicit camera modeling and information-preserving control mechanism for controllable camera simulation in driving scene. The resulting framework DriveCamSim achieves SOTA visual quality and controllability, while unleashing the spatial and temporal-level generalization capability, enabling flexiable camera simulation for downstream application. We hope that DriveCamSim can inspire the community to rethink physically-grounded camera modeling paradigms for driving simulation.

\paragraph{Future work.} Although generalize to camera parameters with small perturbation, we found large perturbation like large translation and rotation in $x$ and $z$ axis result in poor generation result. We leave these problems for future exploration.
\newpage
\small{
\bibliographystyle{plain}
\bibliography{ref}
}
\newpage
\appendix

\section{More implementation Details} 
\label{app:impl}
We train all parameters on 16 RTX 4090 GPUs using AdamW\cite{adamw} optimizer with a linear warm-up of 3000 iterations and learning rate of 2e-4, the total batch size is $16 \times 4 = 64$. The model is trained for 400 epochs in main results and 100 epochs in ablation studies. We only use 2Hz data in nuScenes for training. For building pixel correspondence, we set 10 fixed depth anchor in range of [1, 60] with  linear increasing discretization\cite{center3d}. For overlap-based target view matching, we set the number of target views to twice the number of frames, which is 2 for cross-view attention, and $2 \times (N_r+N_f)$ for reference and temporal attention. For random frame sampling, we randomly sample 4 frames (3 for reference and historical frames and 1 for generation frame) within 12 consecutive frames.

\section{Spatial-level and tempoarl-level generalization} 
\label{app:spatial}
As illustrated before, we choose SparseDrive\cite{sparsedrive} to validate the camera parameter generalization ability of our generative framework. Specifically, SparseDrive is trained with random resize, crop and rotate, making it inherently generalize to slight camera parameter perturbation. Thus, we randomly perturb camera parameters of nuScenes validation dataset, and use a generative model to generate a new dataset called nuScenes-Perturb. The random perturbation is consistent with training augmentation of SparseDrive, so we can compare the metric difference between nuScenes and nuScenes-Perturb to validate the generalization ability of the generative model. As shown in Tab. \ref{tab:spatial_generalization}, our control mechanism not only exhibits best performance on original nuScenes dataset, but also maintains high controllability on nuScenes-Perturb dataset, indicating the model's generalization ability across different camera parameters. We provide more visualization results for spatial-level and temporal-level generalization.

\begin{table}[tb!]
\centering
\caption{Metric difference between original nuScenes dataset and generated nuScenes-Pertube with slight perturbation on camera parameters. Smaller difference indicates model's robust generalization ability on camera parameter variations. "-P" means the metric on nuScenes-Perturb dataset.}
\label{tab:spatial_generalization}
\scriptsize
{    
\resizebox{\linewidth}{!}{
\begin{tabular}{l|c|cc|c|c|cc}
\toprule
\multirow{2}{*}{ID} &
\multirow{1}{*}{Control} &
\multicolumn{2}{c|}{3DOD} & 
\multicolumn{1}{c|}{Tracking} & 
\multicolumn{1}{c|}{Online Mapping} &
\multicolumn{2}{c}{Planning} \\
 & Mechanism& NDS/NDS-P$\uparrow$ & mAP/mAP-P$\uparrow$ & AMOTA/AMOTA-P$\uparrow$ & mAP/mAP-P$\uparrow$ & L2/L2-P$\downarrow$ & Col/Col-P(\%)$\downarrow$ \\
\midrule
1  & Ours & \textbf{36.44}/\textbf{35.80} & \textbf{19.57}/\textbf{19.08} & \textbf{9.05}/\textbf{9.47} & \textbf{22.84}/\textbf{22.02} & \textbf{0.69}/\textbf{0.98} & \textbf{0.19}/\textbf{0.42}  \\
2  & Perspective-based & 31.15/30.11 & 14.26/13.54 & 5.52/5.30 & 21.36/18.18 & 0.73/1.04 & 0.20/0.46  \\
3  & Attention-based & 26.23/25.04 & 10.01/8.96 & 4.92/3.74 & 15.65/13.92 & 0.81/1.05 & 0.32/0.43  \\
\bottomrule
\end{tabular}
}
}
\end{table}

\begin{figure}[htbp]
  \centering
  \includegraphics[width=0.9\linewidth]{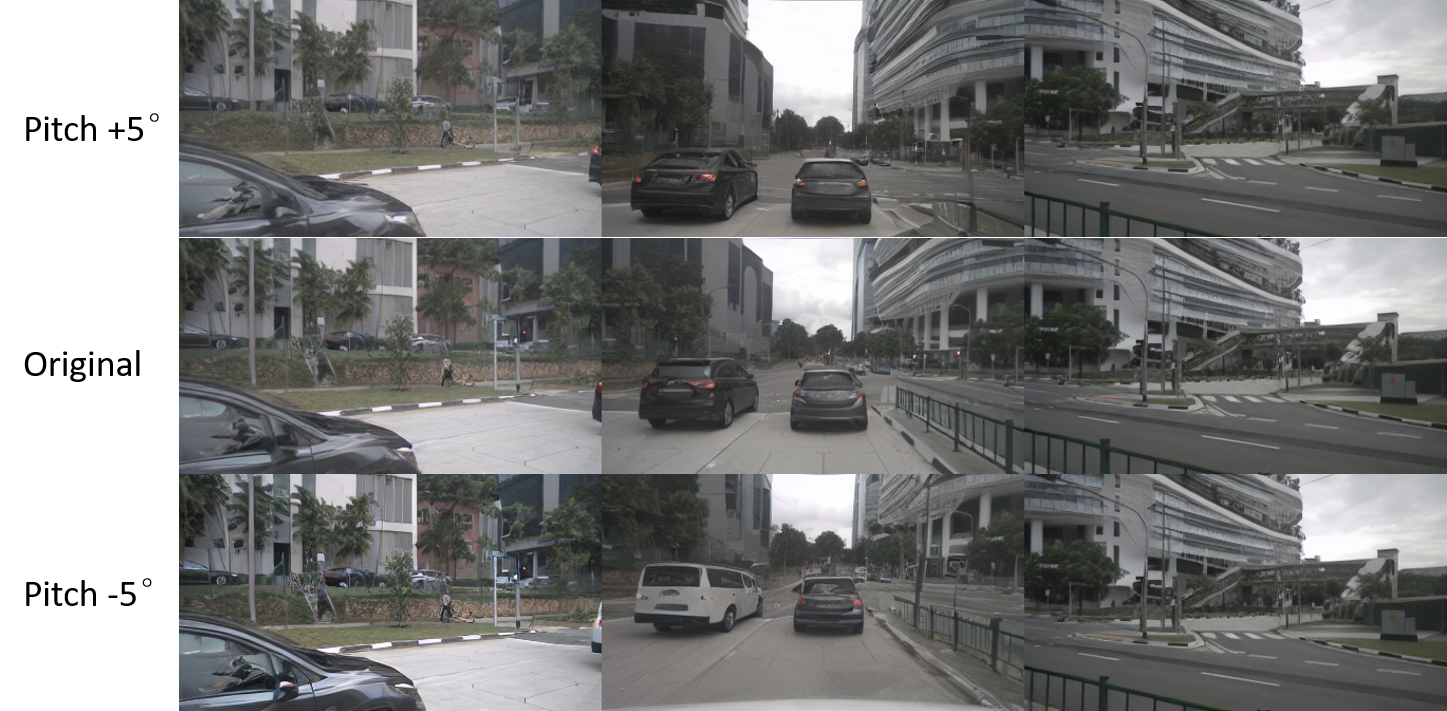}
  \caption{Visualization results for rotating front camera along x-axis.}
  \label{fig:rot_x}
\end{figure}
\begin{figure}[htbp]
  \centering
  \includegraphics[width=0.9\linewidth]{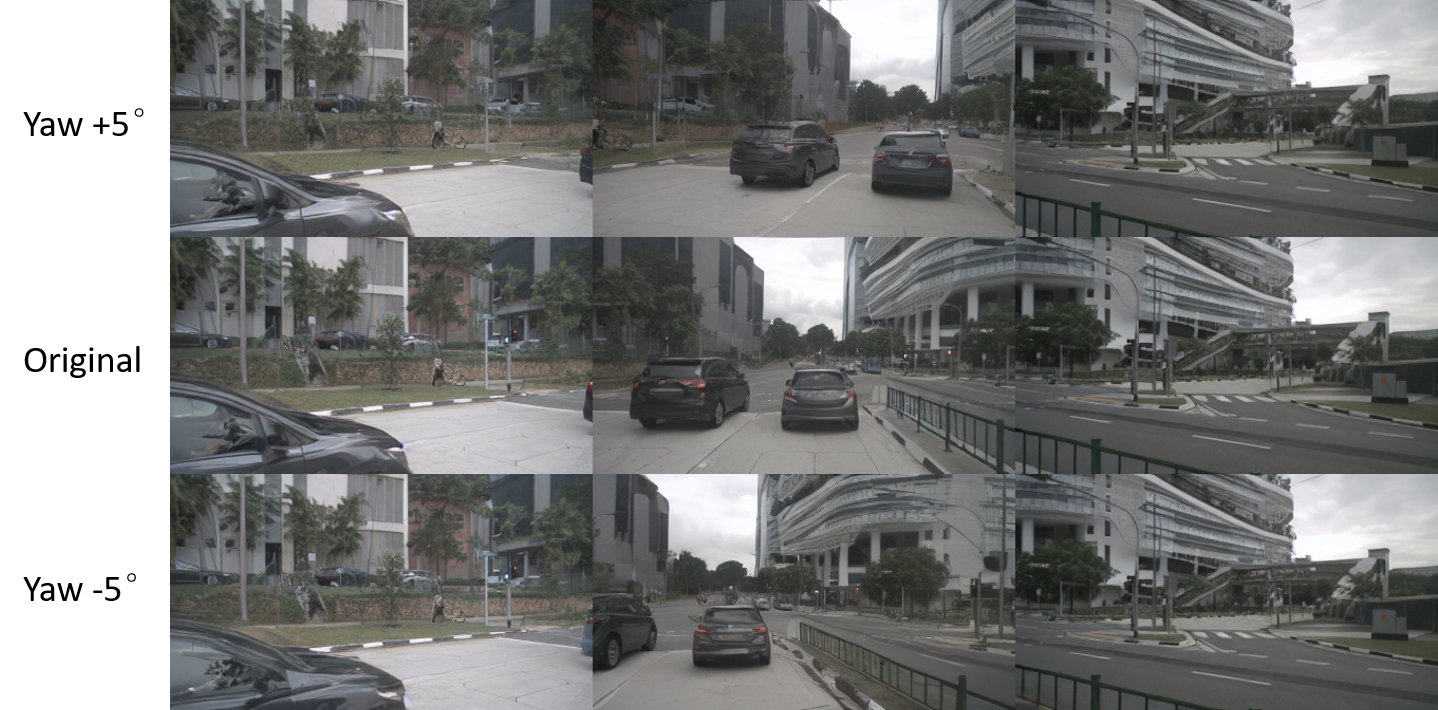}
  \caption{Visualization results for rotating front camera along y-axis.}
  \label{fig:rot_y}
\end{figure}
\begin{figure}[htbp]
  \centering
  \includegraphics[width=0.9\linewidth]{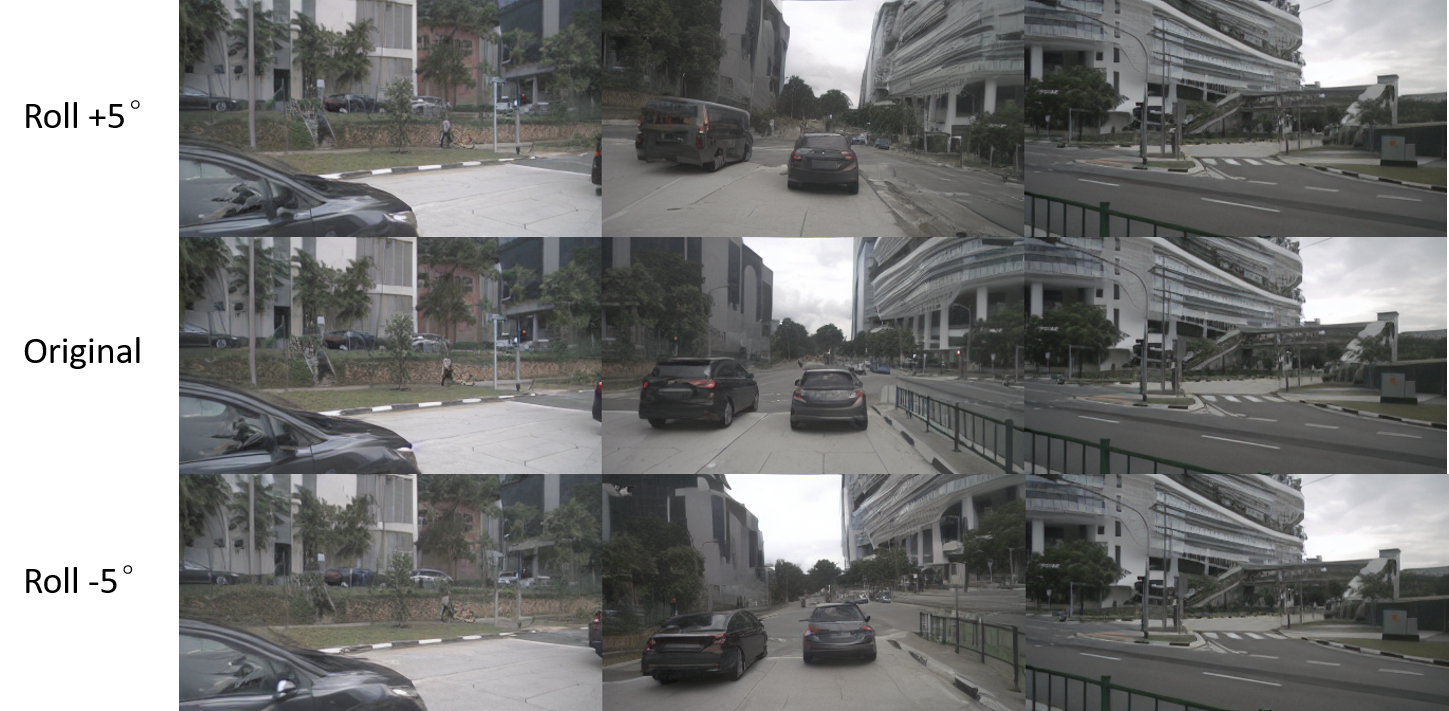}
  \caption{Visualization results for rotating front camera along z-axis.}
  \label{fig:rot_z}
\end{figure}
\begin{figure}[htbp]
  \centering
  \includegraphics[width=0.9\linewidth]{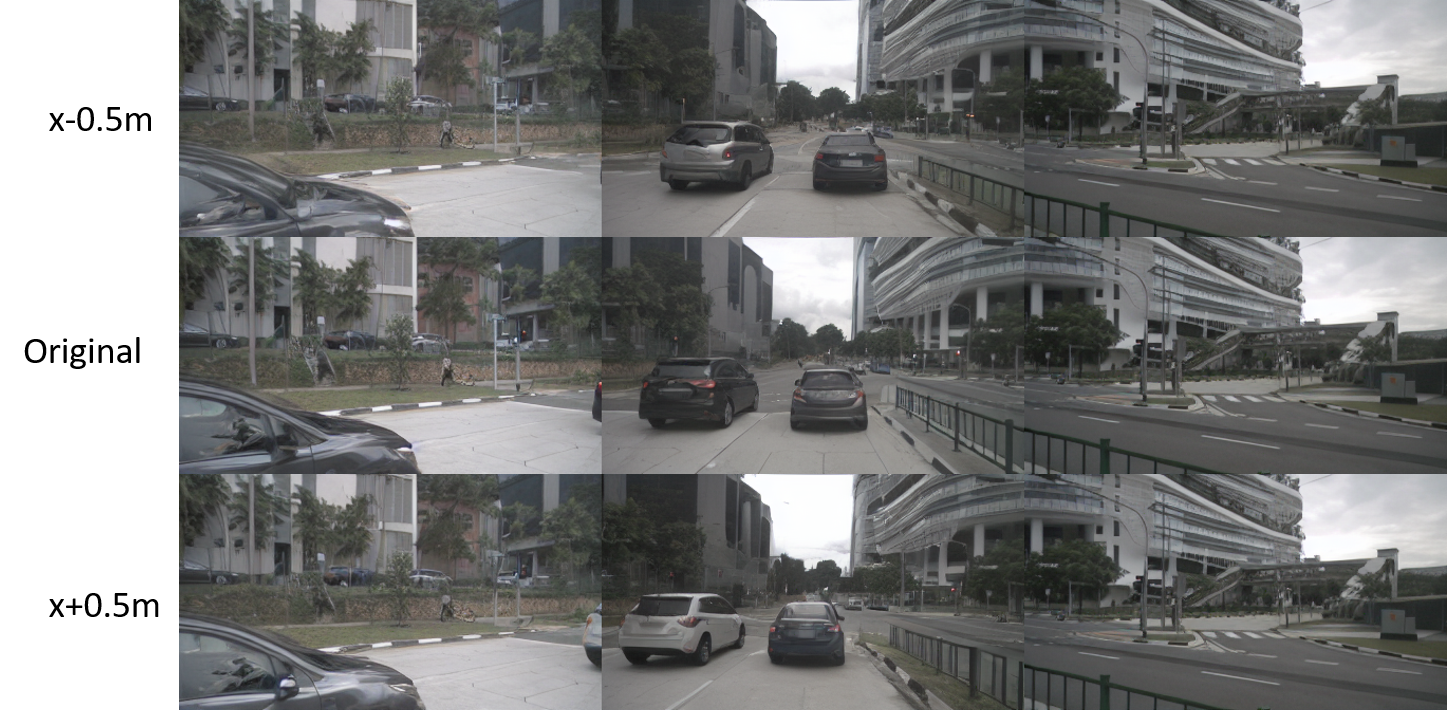}
  \caption{Visualization results for translating front camera along x-axis.}
  \label{fig:trans_x}
\end{figure}
\begin{figure}[htbp]
  \centering
  \includegraphics[width=0.9\linewidth]{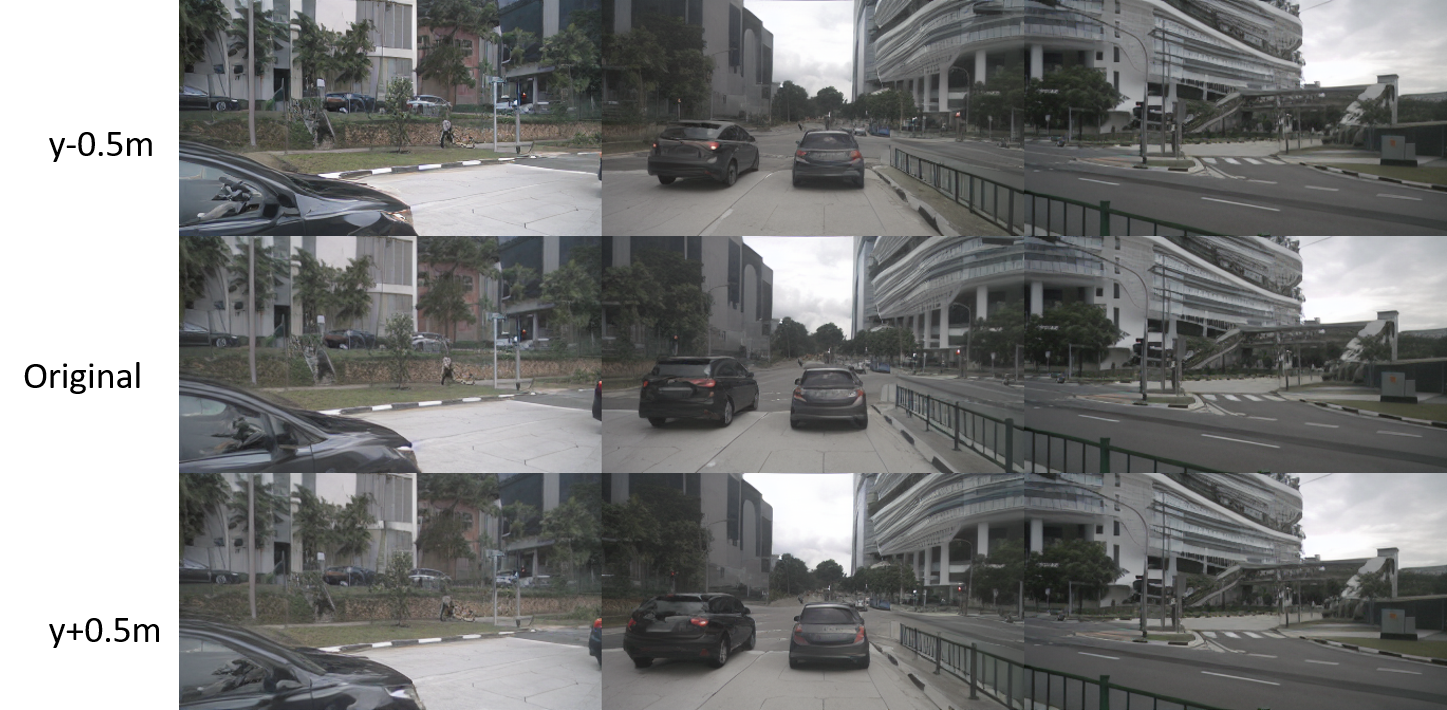}
  \caption{Visualization results for translating front camera along y-axis.}
  \label{fig:trans_y}
\end{figure}
\begin{figure}[htbp]
  \centering
  \includegraphics[width=0.9\linewidth]{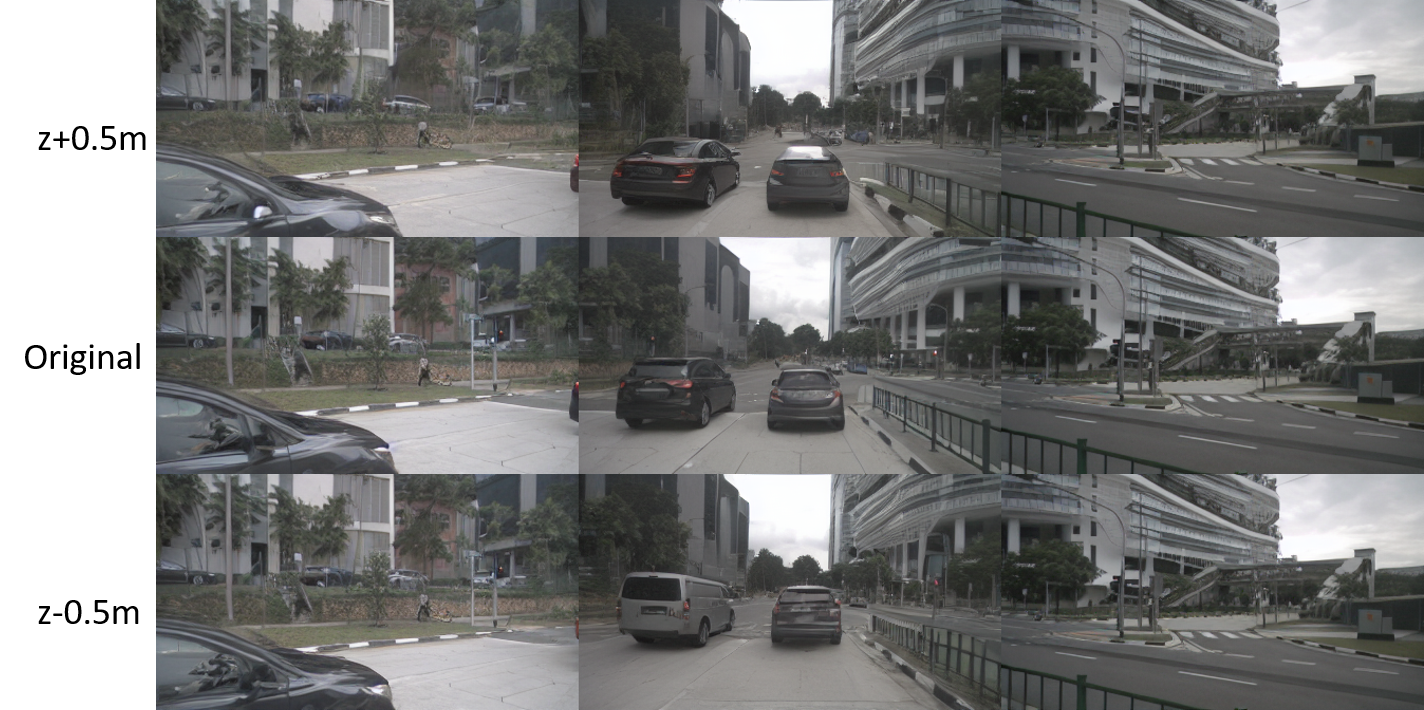}
  \caption{Visualization results for translating front camera along z-axis.}
  \label{fig:trans_z}
\end{figure}
\begin{figure}[htbp]
  \centering
  \includegraphics[width=0.9\linewidth]{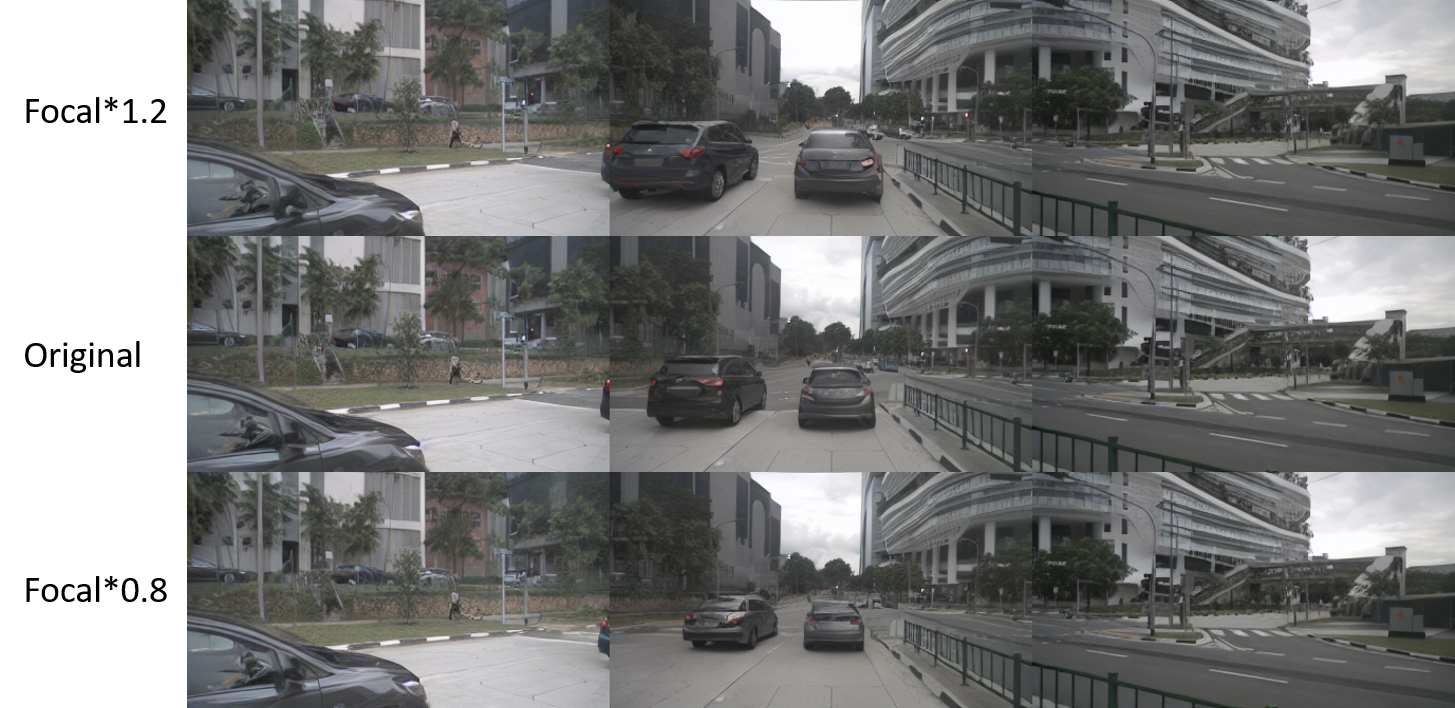}
  \caption{Visualization results for scaling focal length.}
  \label{fig:intrinsic}
\end{figure}
\begin{figure}[htbp]
  \centering
  \includegraphics[width=0.9\linewidth]{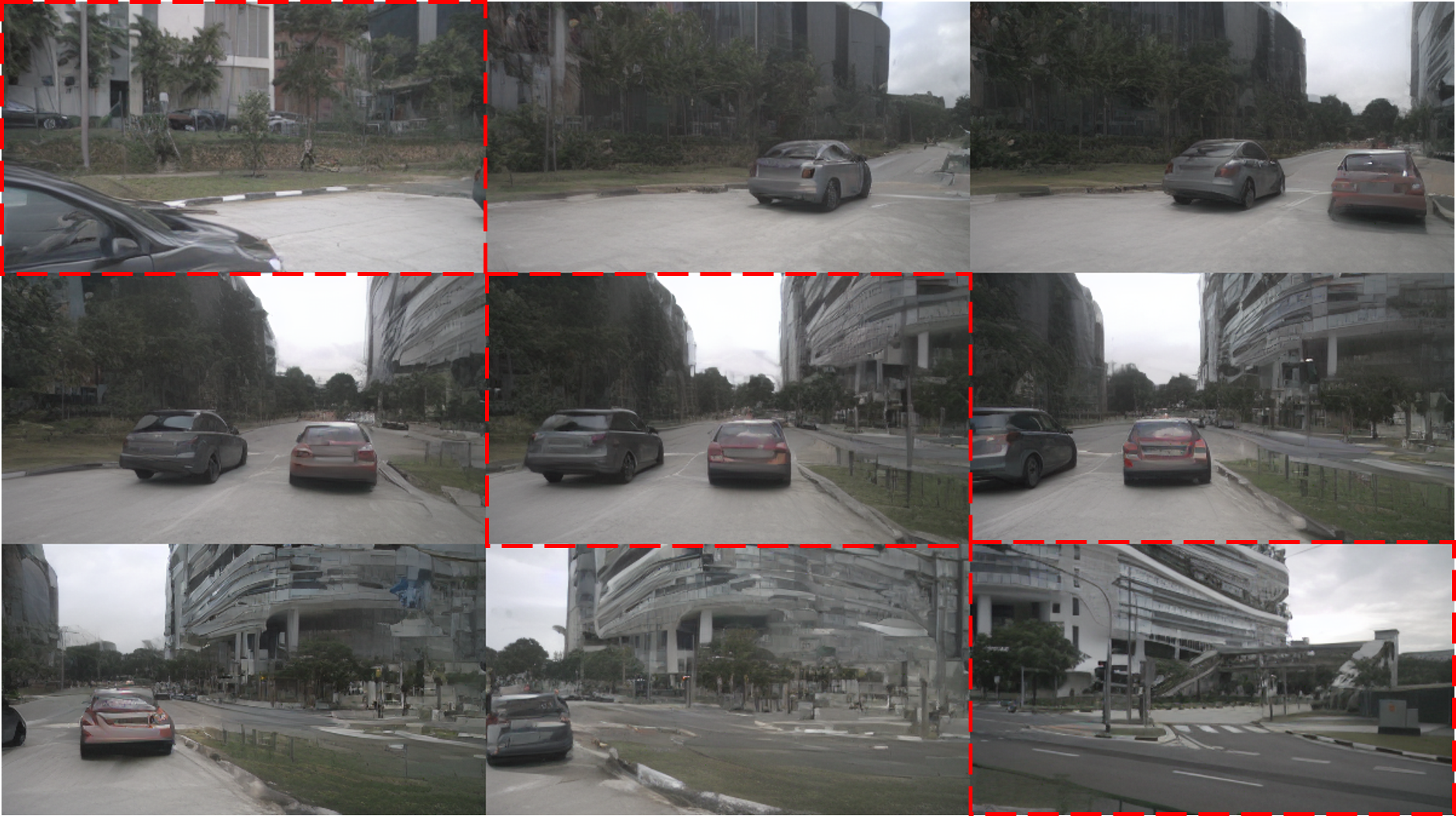}
  \caption{Visualization results for inserting 3 virtual cameras on both sides of the front camera with different yaw angle. 3 views with red border are front-left camera, front camera and front-right camera of nuScenes dataset, while others are virtual cameras.}
  \label{fig:virtual_cam}
\end{figure}
\begin{figure}[htbp]
  \centering
  \includegraphics[width=\linewidth]{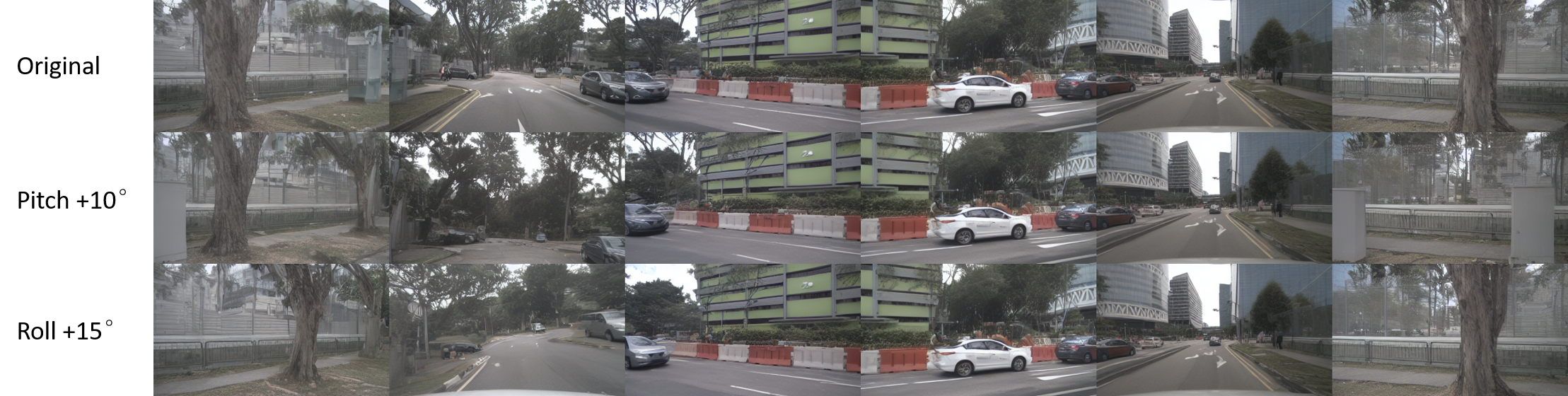}
  \caption{Failure cases of large rotation of front camera.}
  \label{fig:rot_z}
\end{figure}
\begin{figure}[htbp]
  \centering
  \includegraphics[width=\linewidth]{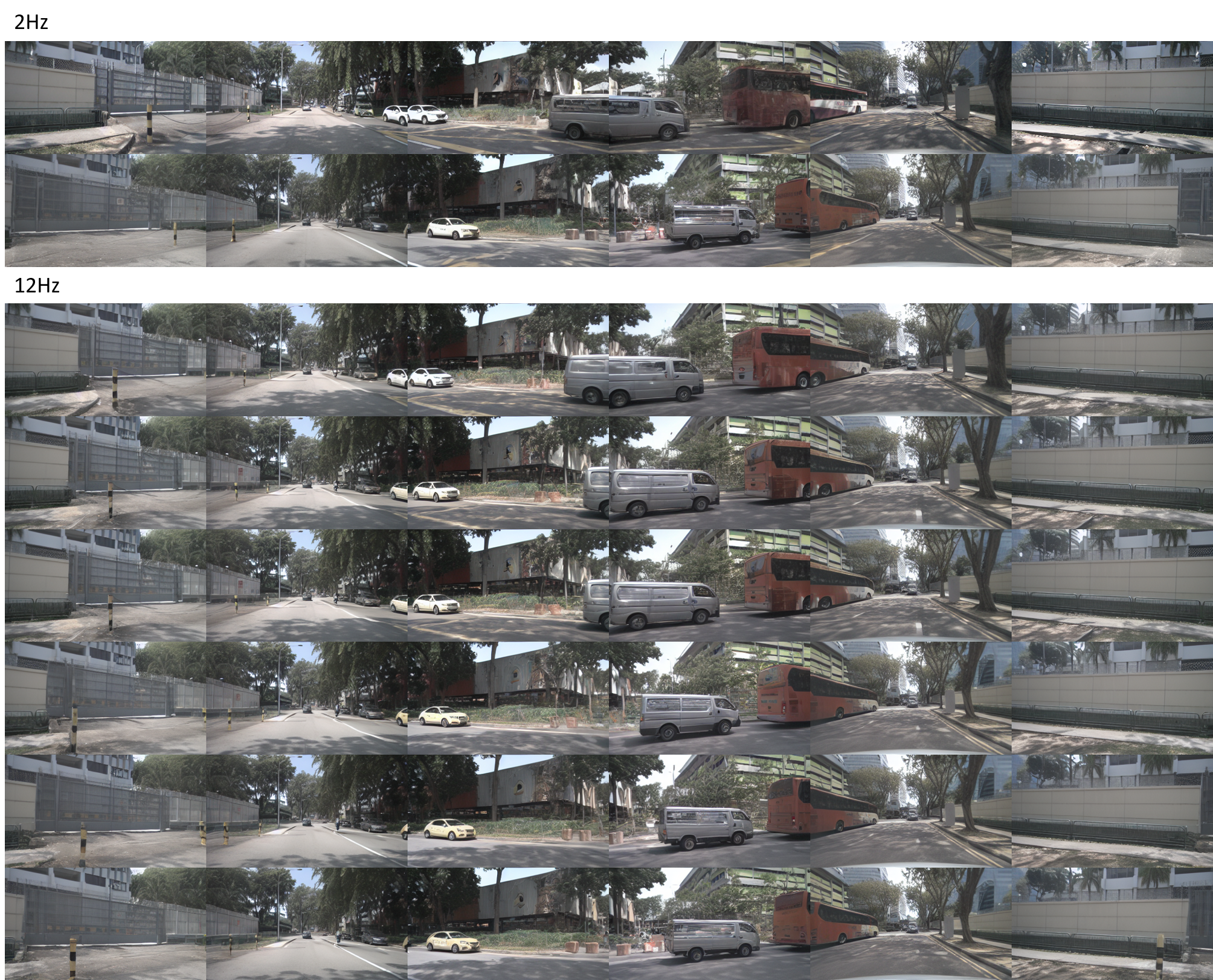}
  \caption{Qualitative results for temporal-level generalization. Trained on 2Hz data, our model can generalize to high-frequency 12Hz generation.}
  \label{fig:temporal_generalization}
\end{figure}
\begin{figure}[htbp]
  \centering
  \includegraphics[width=\linewidth]{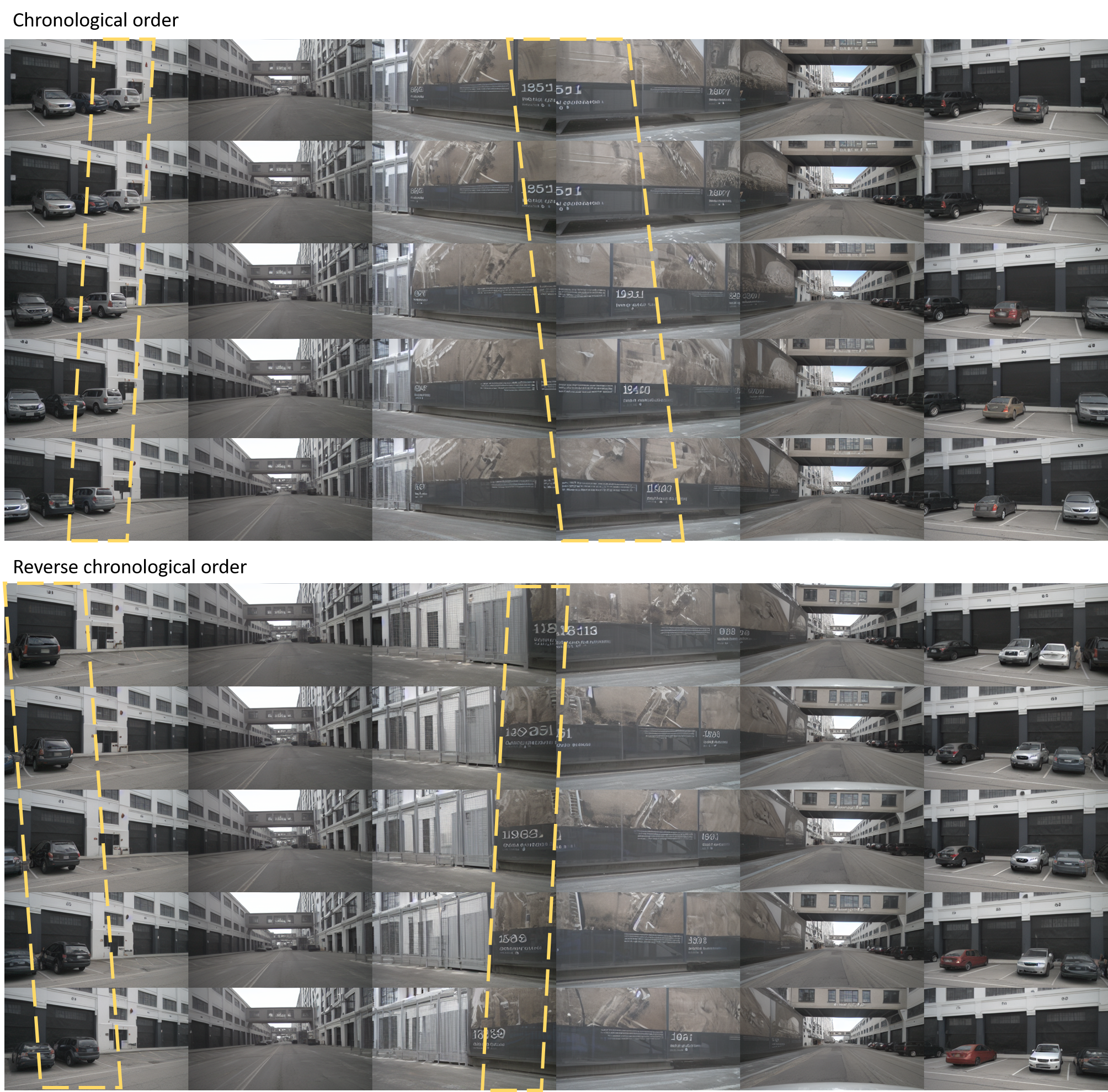}
  \caption{Qualitative results for temporal-level generalization. Our model can generate videos in reverse chronological order to simulate the scene that ego vehicle is moving backward.}
  \label{fig:reverse}
\end{figure}
\begin{figure}[htbp]
  \centering
  \includegraphics[width=\linewidth]{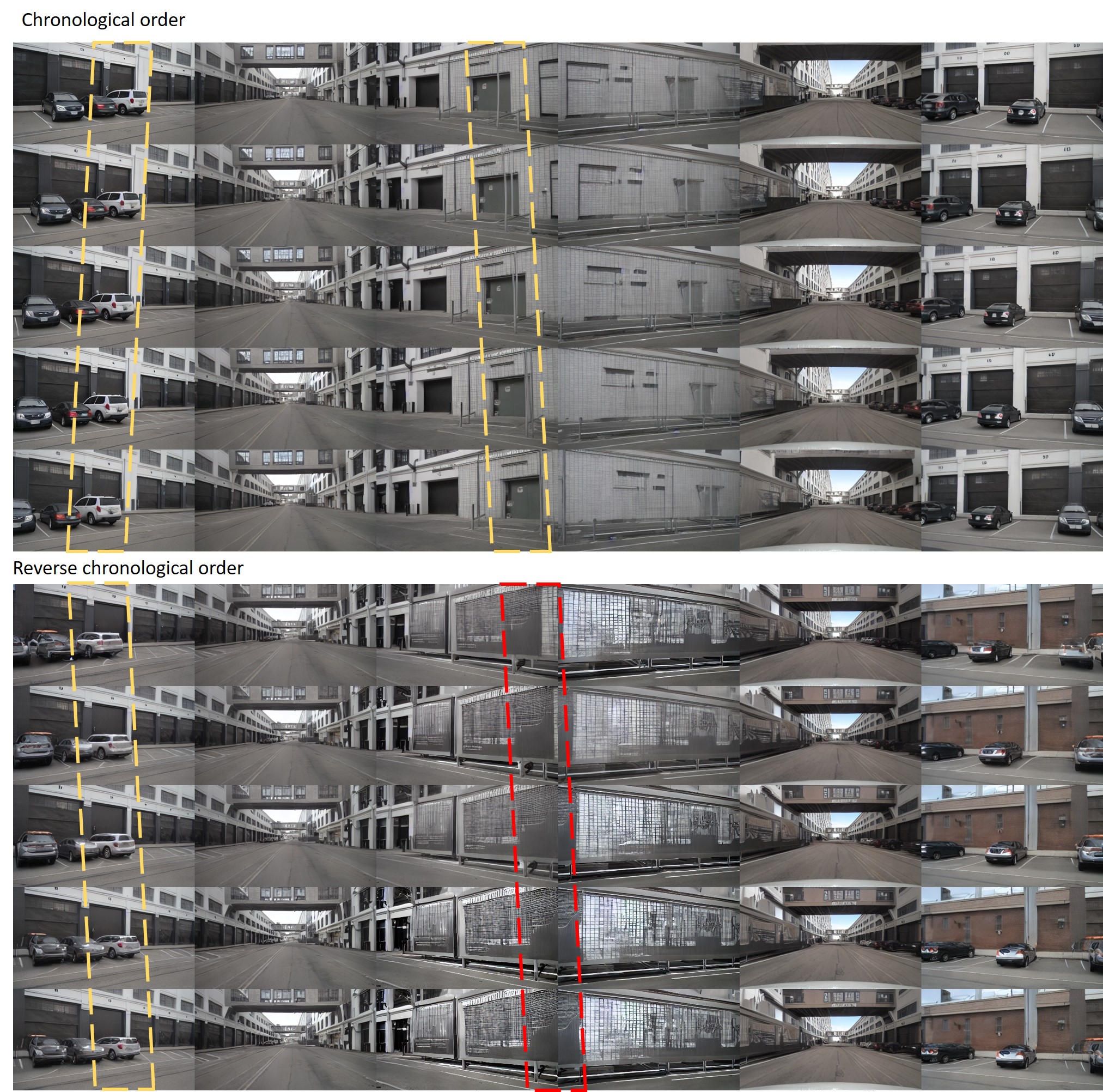}
  \caption{Qualitative results for temporal-level generalization of baseline model DreamForge\cite{dreamforge}. When generating in chronological order, the foreground and background should move forward relative to ego vehicle. DreamForge can generate foreground objects correctly due to perspective-based control, but cannot handle background correctly with implicit camera modeling.}
  \label{fig:reverse}
\end{figure}

\end{document}